\documentclass[conference]{IEEEtran}
\IEEEoverridecommandlockouts
\usepackage{cite}
\usepackage{amsmath,amssymb,amsfonts}
\usepackage{algorithmic}
\usepackage{graphicx}
\usepackage{textcomp}
\usepackage{xcolor}
\usepackage{subcaption}
\usepackage{multirow}
\usepackage{float}
\usepackage{hyperref}
\def\BibTeX{{\rm B\kern-.05em{\sc i\kern-.025em b}\kern-.08em
    T\kern-.1667em\lower.7ex\hbox{E}\kern-.125emX}}
\begin{document}

\title{Memory Guided Road Detection}

\author{\IEEEauthorblockN{Praveen Venkatesh*}
\IEEEauthorblockA{\textit{Electrical Engineering} \\
\textit{Indian Institute of Technology}\\
Gandhinagar, India }
\and
\IEEEauthorblockN{Rwik Rana*}
\IEEEauthorblockA{\textit{Mechanical Engineering} \\
\textit{Indian Institute of Technology}\\
Gandhinagar, India}
\and
\IEEEauthorblockN{Varun Jain*}
\IEEEauthorblockA{\textit{Electrical Engineering} \\
\textit{Indian Institute of Technology}\\
Gandhinagar, India }
}

\maketitle

\begin{abstract}
In self driving car applications, there is a requirement to predict the location of the lane given an input RGB front facing image. In this paper, we propose an architecture that allows us to increase the speed and robustness of road detection without a large hit in accuracy by introducing an underlying shared feature space that is propagated over time, which serves as a flowing dynamic memory. By utilizing the gist of previous frames, we train the network to predict the current road with a greater accuracy and lesser deviation from previous frames. 
\end{abstract}
% \section{TODO}
% Title of the project
% (ii) Problem Statement
% (iii) Individual Contribution in the project and link to code (if any)
% (iv) Approach (with block schematic if applicable)
% (v) Quantitative and Qualitative Results
% (vi) Your key observations
% (vii) Describe your contribution: Is it novel? What modifications are done to the existing
% method? Is it a direct implementation of the exiting method?
\section{Introduction}
Tracking roads is a very important task in self-driving vehicles, where it is important to know the location of the roads, and where the car is positioned with respect to the road. Several existing methods rely on either algorithmic models of the roads, or using segmentation based models for predicting roads like integrating RANSAC with CNNs, RBMs, CRFs etc. However, these methods fail to address the temporal consistency of the act of road detection, and treat the problem as independently predicting roads for each frame.

\begin{figure}[H]
    \centering
    \vspace{-0.3cm}
    \includegraphics[width = 0.9\linewidth]{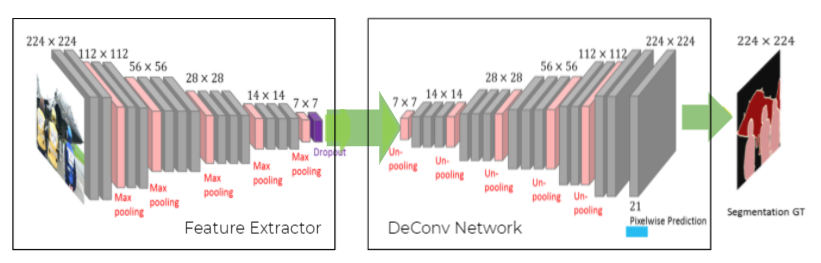}
    \caption{FCN for Segmentation\cite{kim}}
    \label{FCNSeg}
\end{figure}

In this paper, we use a combination of large and small feature extractors that learn to infer from a shared underlying flowing memory allowing for fast inference with a low accuracy dip. The proposed method takes inspiration from the human eye as to how a human can perceive complex scenes and infer a rich representation of the scenes, based on previously seen scenes.

\begin{figure}[H]
    \centering
    \vspace{-0.5cm}
    \includegraphics[width = 0.5\linewidth]{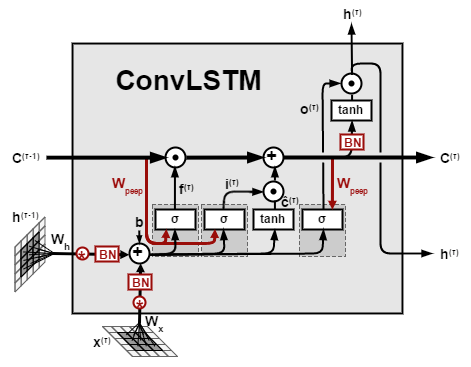}
    \caption{Conv LSTM Model\cite{xavier_2019}}
    \label{ConvLstm}
\end{figure}

Convolutional LSTMs (Long Short Term Memory) can be used to remember and track images instead of applying object detection techniques will help save time, computational power and will be more efficient. Thus an LSTM, based memory model is proposed to detect an object in a single frame and then tracking the object using and hybrid of Resnet 18, Resnet 101, FCN 32 and Conv LSTM network.
Thus this model makes the process computationally fast and is robust and thus can also be deployed on mobile devices for a smooth application of road detection. We primarily take inspiration from the paper Looking Fast and Slow: Memory-Guided Mobile Video Object Detection, where Liu et al\cite{liu_looking_2019} propose a method for fast object detection by using different capacity models.

\section{Related Works}
With the onset of machine learning, deep learning and computer vision techniques, the area of semantic segmentation has been increasingly studied. Early works such as those introduced by Long et al(2014) \cite{shelhamer_fully_2017} are called FCN's (Fully Convolutional Networks). The main aim of the FCN was to replace fully connected layer by fully convolutional layers. This allows for a large reduction in the number of parameters while also introducing spatial consistency over localized patches of an image. Fisher et al(2015) \cite{yu_multi-scale_2016} suggested a dilated convolutional method whose is based on the exponential expansion of the receptive field without loss of resolution or coverage. 
\begin{figure}[H]
    \centering
    \includegraphics[width=0.5\linewidth]{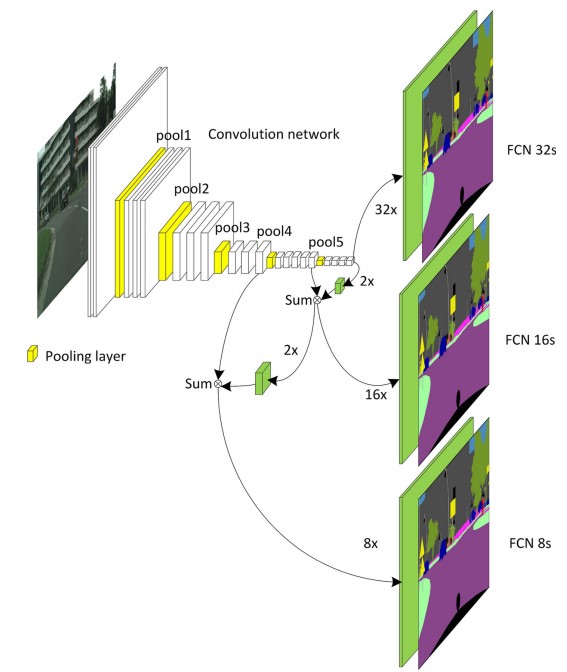}
    \caption{A schematic of FCN implemented by Long et al}
    \label{long fcn}
\end{figure}

Given the volume of work done in this area, a large number of datasets such as the Microsoft COCO, PASCAL, ImageNet, SUN, CityScape have been created\cite{deng2009imagenet}, \cite{noauthor_pascal_nodate},\cite{noauthor_microsoft_nodate}. Moreover several simulators such as the CARLA, LGSVL, DeepDrive, AirSim etc.\cite{dosovitskiy_carla_2017}, \cite{rong_lgsvl_2020}, \cite{shah_airsim_2017} have been extensively used to test autonomous agents safely using synthetic data. They include various functionalities ranging from semantic segmentation to depth data to help train algorithms for autonomous vehicles.

\begin{figure}[H]
    \centering
    \includegraphics[width=0.9\linewidth]{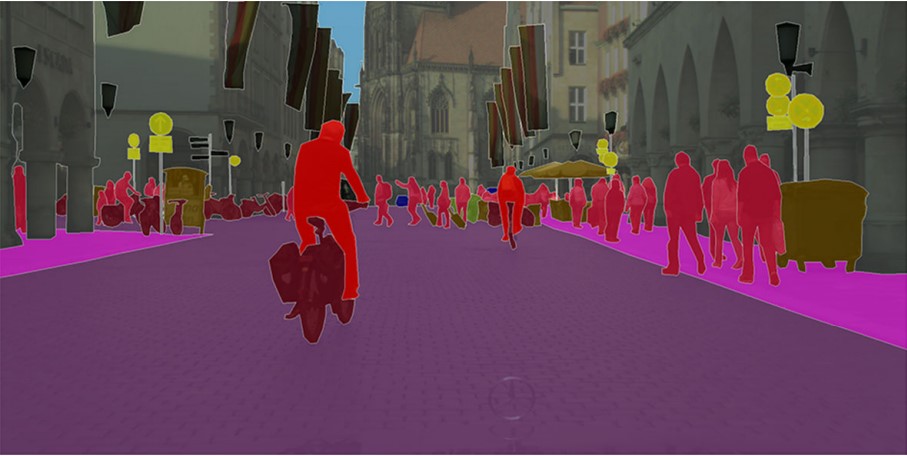}
    \caption{A image from the Cityscape Dataset}
    \label{fig:my_label}
\end{figure}

\section{Dataset}

For our experiments, we use a dataset collected manually from the CARLA simulator. The dataset along with their splits can be found here: \href{https://drive.google.com/drive/u/1/folders/1lNtkBCYgI6VG5arNmLh1cVMz128XWzUT}{\textit{Link}}.
The dataset is collected using the Town01 map, with the weather settings set to rainy. We collect the front facing RGB images, and the corresponding road segmentation using the available segmentation maps.
In order to speed up the training process, we store the dataset in a single file as a list of images that is loaded entirely into memory during training. We also double the size of the dataset collected by using the flipped versions of the road and RGB images as inputs.

\section{Our Approach}

Our approach consists of three main cascaded stages - 
\begin{enumerate}
    \item Feature Extraction
    \item Shared Memory
    \item Decoder
\end{enumerate}

The memory module and the segmentation generator are shared between multiple models, whereas we utilize two feature extractors for our experiments.

\subsection{Feature Extractors - \textit{The Interpreters}}

In our experiments, we use primarily use 2 variants of pretrained ResNet based architectures. Since the goal of the method is to maintain accuracy while increasing speed, we use 2 drastically different sized variants, namely the ResNet-18 and the ResNet-101 variants. The ResNet is used as a low capacity feature extractor that runs at almost 3x the speed of the ResNet-101 which is slower, but performs better. We use the weights downloaded from the Torch Model zoo which has been pretrained on the ImageNet dataset for classification tasks.

Since there are only 2 classes (no road / road), we use binary labelling where road is marked as 1 and no road is marked as 0. We use the Adam optimizer along with binary cross entropy loss for segmentation.

% While making the model, both pretrained models of Resnet and VGG were used as the main interpreter for the input images. These models were used to classify the input images into two classes namely 0 - No Road, 1- Road. The entire model showed more accuracy and speed on the ResNet block, namely because of their architecture. 

Initially, we perform our experiments on two VGG variants (VGG11 and VGG16). However, VGG models have a significantly larger number of parameters leading to lower speeds. However, we later switch to the two ResNet variants described above due to their ability to produce extremely high accuracy in classification techniques using considerably less training parameters. For example \cite{kolesnikov_big_2020}, produces an accuracy of 98.4\%.

% Resnet18 is used for its high speed inference and Resnet101 is used because of its high level of accuracy. Basically, Resnet18 weights are trained to imitate the Resnet101 by calculating the loss between the Resnet 18 and trained Resnet101 outputs rather than Resnet18 and ground truth. This ensures that both the speed of Resnet18 and accuracy of Resnet101 are taken into account.

% Pretrained Resnet18 and Resnet101 upto the second last layer and is combined with a 2D convolutional layer give a 512 channel output, which becomes an input to the temporal module. 

To extract features from the ResNet models, we remove the last classification layers, and use the intermediate feature vector that is generated (512 and 2048 dimensions for ResNet18 and ResNet 101). To maintain the same dimension of the feature vector, we add convolutional layers to reduce the dimensionality of the features to 512 dimensions for both the ResNet variants. The following images are the results corresponding to only Resnet18, Resnet101 and interleaving of Resnet18 \& Resnet101.

\subsection{Shared Memory - \textit{The Time Keeper}}

\begin{figure}[h]
    \centering
    \includegraphics[scale = 0.4]{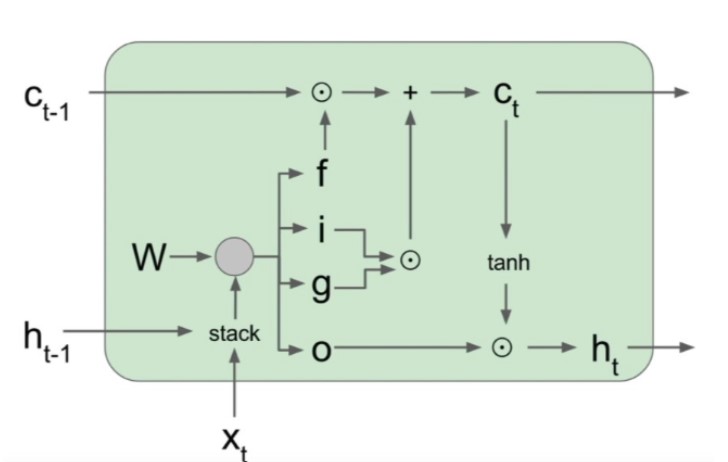}
    \caption{A LSTM cell}
    \label{lstm cell}
\end{figure}

The main aim of the model is to exploit the temporal consistency of time series data along with spatial data. A natural fit is hence the concept of the convolutional LSTMs, which is essentially an LSTM cell which operates on convolutional gates instead of connected gates. The LSTM stores the $h_{i-1}$ hidden state and inputs the hidden states along with the ResNet input in the $i^{th}$ time step. The 512 channel output of the ResNet block, becomes the input to the convolutional LSTM block. Our model employs a single LSTM cell (Fig : \ref{lstm cell}). The LSTM cell provides a 128 channel output $C_i$ along with a cell state $h_i, c_i$ which is the hidden state and current state of the ConvLSTM of the $i^{th}$ timestep. The ConvLSTM is implemented as a stateful LSTM where the hidden states are propogated through frames, which thus giving a temporal feature to the enitre process.

\subsection{Generating Segmentations - \textit{The Decoder}}
\begin{figure}[H]
    \vspace{-1cm}
    \centering
    \includegraphics[width=\linewidth,height=4cm]{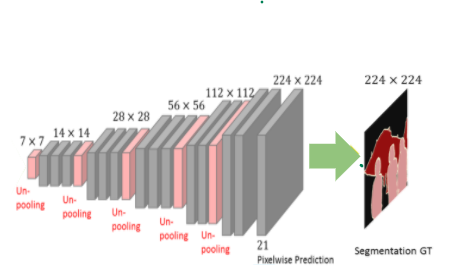}
    \caption{Decoder for FCN32s}
    \label{FCN32}
\end{figure}

Once the information has passed through the ConvLSTM, the outputs of the LSTM model are passed through a decoder block which gives a segmented image of the inferred road. The decoder block is an up-sampling fully convolutional network made up of 5 transpose convolutional layers as shown in Fig.\ref{FCN32}. The up-sampling layers in later joined with a convolutional layer to predict the final road mask. The FCN32 decreases the channels obtained from the LSTM from 128 to 3 and up-samples the image to the required size.  

\subsection{Pipeline}
\begin{figure}[H]
    \centering
    \includegraphics[width = 0.95\linewidth]{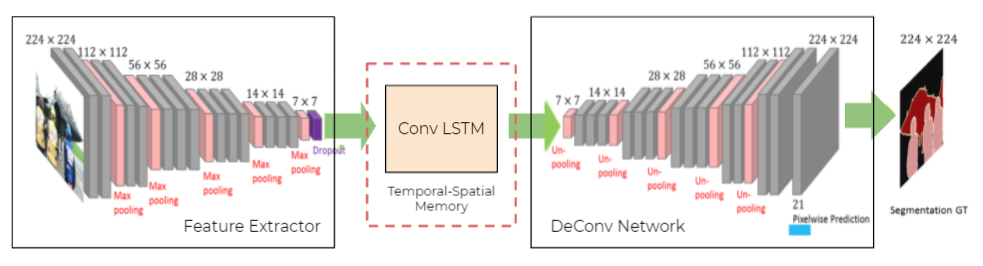}
    \caption{Model with the Temporal-Spatial Memory}
    \label{fcnlstm}
\end{figure}

Integrating all of the parts, we show that 3 stages are cascaded 
\begin{itemize}
    \item Feature Extraction -The Interpreter
    \item Shared Memory - The Time keeper
    \item Segmentation Generation - The Decoder
\end{itemize}
The block diagram of the architecture is shown in Figure \ref{fcnlstm}.

\section{Interleaving - Training Pipeline}

In order to train the ConvLSTM and the deconvolutional network, we use the following approach:
\begin{itemize}
    \item At the beginning of a sequence of 6 images, randomly choose a feature extractor and pass the image through it.
    \item Pass the extracted features through the ConvLSTM and the Deconv block.
    \item Randomly sample a feature extractor and pass the next frame through it.
    \item Pass the previous hidden states along with the current feature through the LSTM and Deconv block.
    \item Repeat for the sequence of 6 frames.
\end{itemize}
An illustrative diagram of this approach is show in Fig \ref{training pipeline}. We set the extractor selector to choose the ResNet101 extractor more often than the ResNet18 extractor as it has greater accuracy. This is achieved by randomly generating a number and checking if it is greater than a threshold epsilon and thus, a feature extractor is selected.
\begin{figure}[H]
    \centering
    \includegraphics[width=0.9\linewidth]{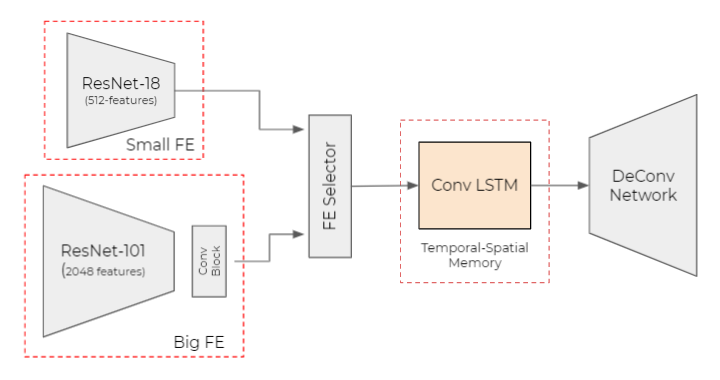}
    \caption{Training Pipeline}
    \label{training pipeline}
\end{figure}

\begin{figure}[H]
    \centering
    \includegraphics[width=0.9\linewidth]{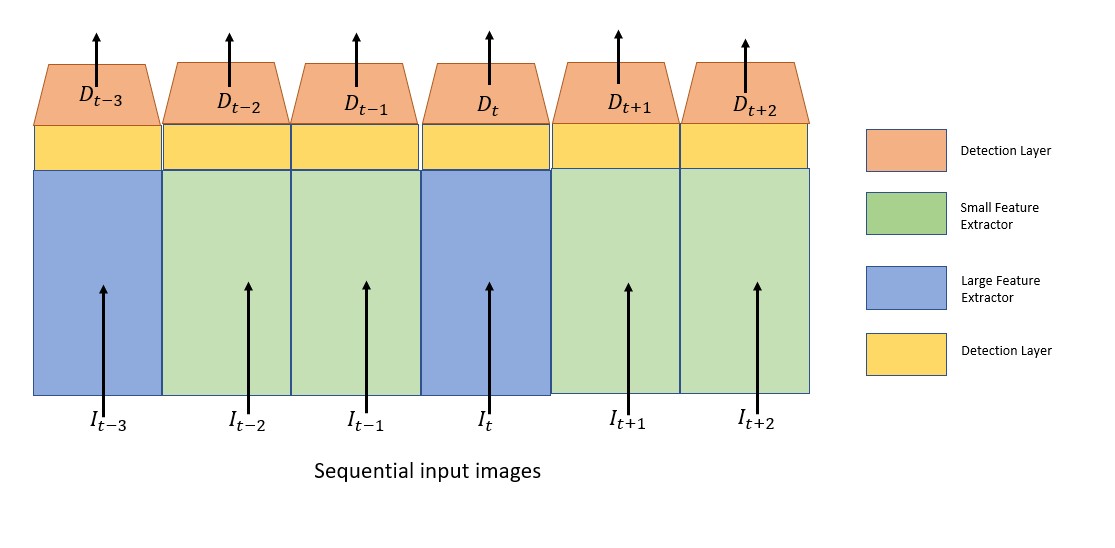}
    \caption{Interleaving}
    \label{training pipeline}
\end{figure}

In our experiments, we utilize two different methods for introducing temporal consistency - 
\begin{itemize}
    \item \textbf{Batched Interleaving} - The sequence of 6 images are generated by duplicating the same image 6 times.
    \item \textbf{Sequential Interleaving} - The+ sequence of 6 images are taken from the actual sequence.
\end{itemize}

% \subsection{Batched Interleaving Technique}

% 6 160x160 image copies i.e a Batch are taken and put as an input to the LSTM block. The inputs in the LSTM cell leads to formation of a hidden state and thus the temporal data. These temporal data are used in successive images and thus help in increasing the accuracy of the model. The convolutional LSTM takes a 512 channel input and gives a 512 channel output. The output is then passed through the the deconvolutional block. The deconvolutional block is a series of 5 convolutional transpose layers. This deconvolutional blocks acts as an upsampler for the model and thus generating the output image. The output image is then compared with the road mask i.e. we find the loss using Binary crossentropy. The last step of the updating the weights using back-propagation. An Adam optimizer is used with learning rate as 0.01. 
% We create 2 feature extractors

% using n=6 copies of the images, we pass it through the FE, and LSTM to form the temporal spatial memory. Once the memory is formed, we once again pass the image through the LSTM and then get the seg map using the decoder network. using gt we calc loss using binary crossentropy, then update weights.

% \subsection{Sequential Interleaving Technique}
% Now we take 6 sequential images i.e. 6 6 consecutive frames from a video. This is then passed through the same blocks i.e. the convolutional LSTM and the deconvolutional network to give the final result and thus we sequential interleave the images to get the final output. 

\section{Results}

    % \hspace*{0.1 cm}   % maximizeseparation between the subfigures

For baseline comparisons, we train FCN32s deconvolution block, preceeded by different bare features extractors. We observe that using a small Feature extractor like the Resnet18, gives us a faster inference time trading off our accuracy, while using a more intricate feature extractor like ResNet-101, we get better results at the cost of inference time. 

Now that our baselines are set, we setup our two training pipelines on the various data loading techniques as discussed in the earlier section.

\textbf{Batched \& Interleaved model} is the model that trains over similar images to create the lstm memory. For training the model, we replicate the images and pass them through the feature extractor and the ConvLSTM, to prepare the temporal memory. Using the memory, we infer the last image and calculate the loss after decoding the segmentation map with ground truth.

\textbf{Sequential \& Interleaved model} is the model that trains over 6 sequential images. The LSTM memory is generated using the 5 sequential images. Using this temporal memory we infer the 6th image and calculate the loss. The weights are updated using the obtained loss.

During inference, unlike the training pipeline, we infer models using strategies that are predecided. In our experiments, we try making choices depending on the order (Inference the large model for 1 in n frames, or by using thresholding of a random number generator). In our pipeline, we clear the weights whenever the large feature extractor is run. We do this as the results accumulate over time if the weights are not cleared.

\subsection{Qualitative Results}

\begin{figure}[H]
    \centering
    \vspace{-0.5cm}
      \begin{subfigure}{0.32\linewidth}
        \includegraphics[width=\linewidth,height=3cm]{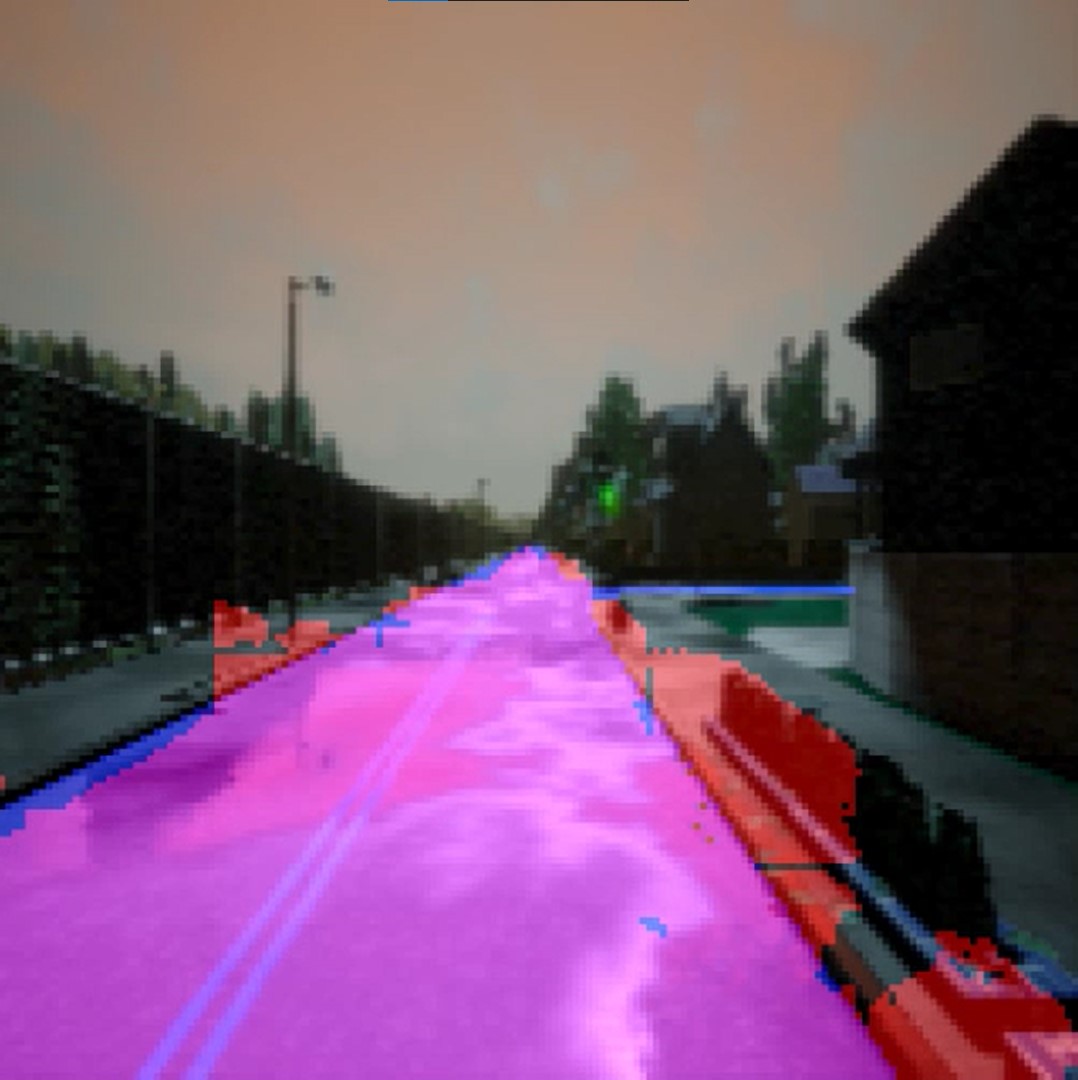}
        \caption{} \label{fig:1a}
      \end{subfigure}%
    \hspace*{0.1 cm}   % maximize separation between the subfigures
      \begin{subfigure}{0.32\linewidth}
        \includegraphics[width=\linewidth,height=3cm]{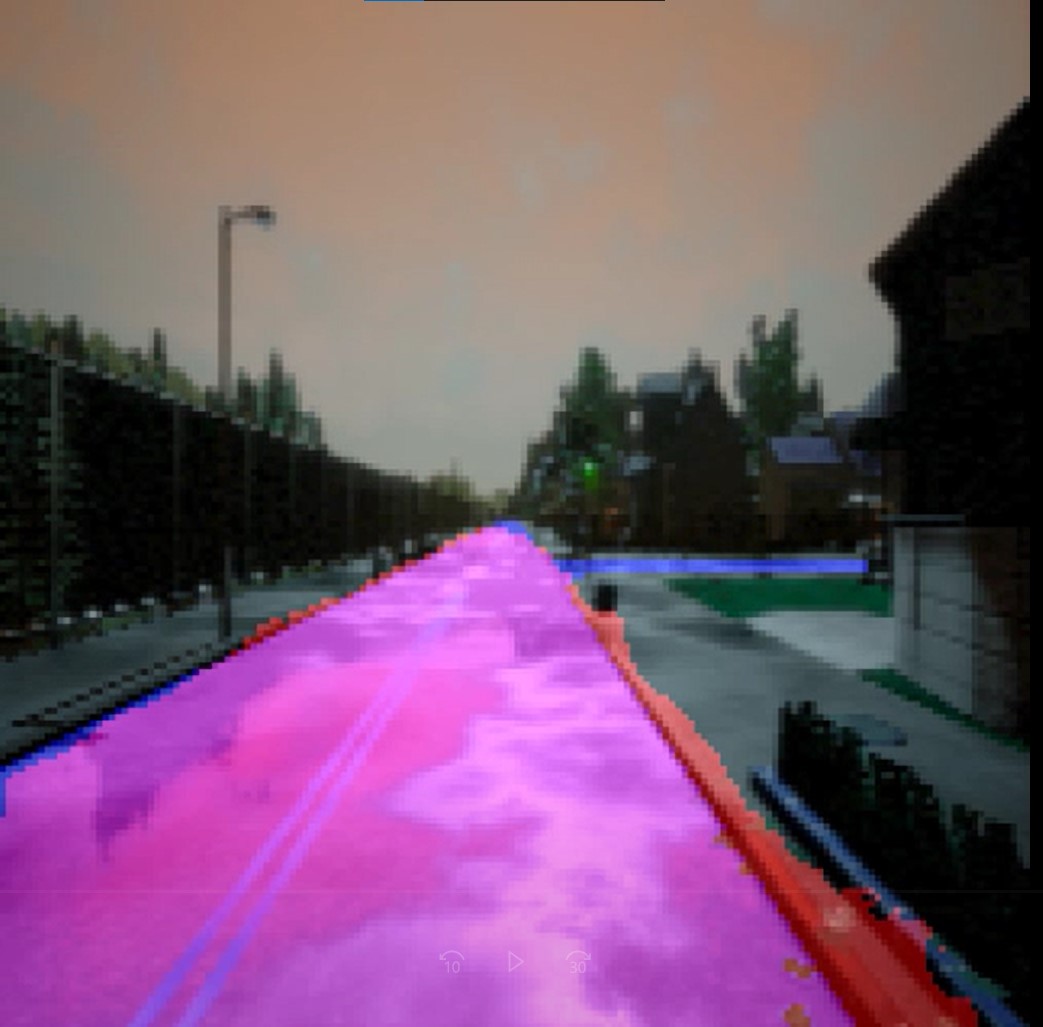}
        \caption{} \label{fig:1b}
      \end{subfigure}
    \caption{(a) Vanilla ResNet18 (b) Vanilla ResNet101} \label{fig:1}
\end{figure}

\begin{figure}[H]
    \centering
    \vspace{-1cm}
      \begin{subfigure}{0.32\linewidth}
        \includegraphics[width=\linewidth ,height=3cm]{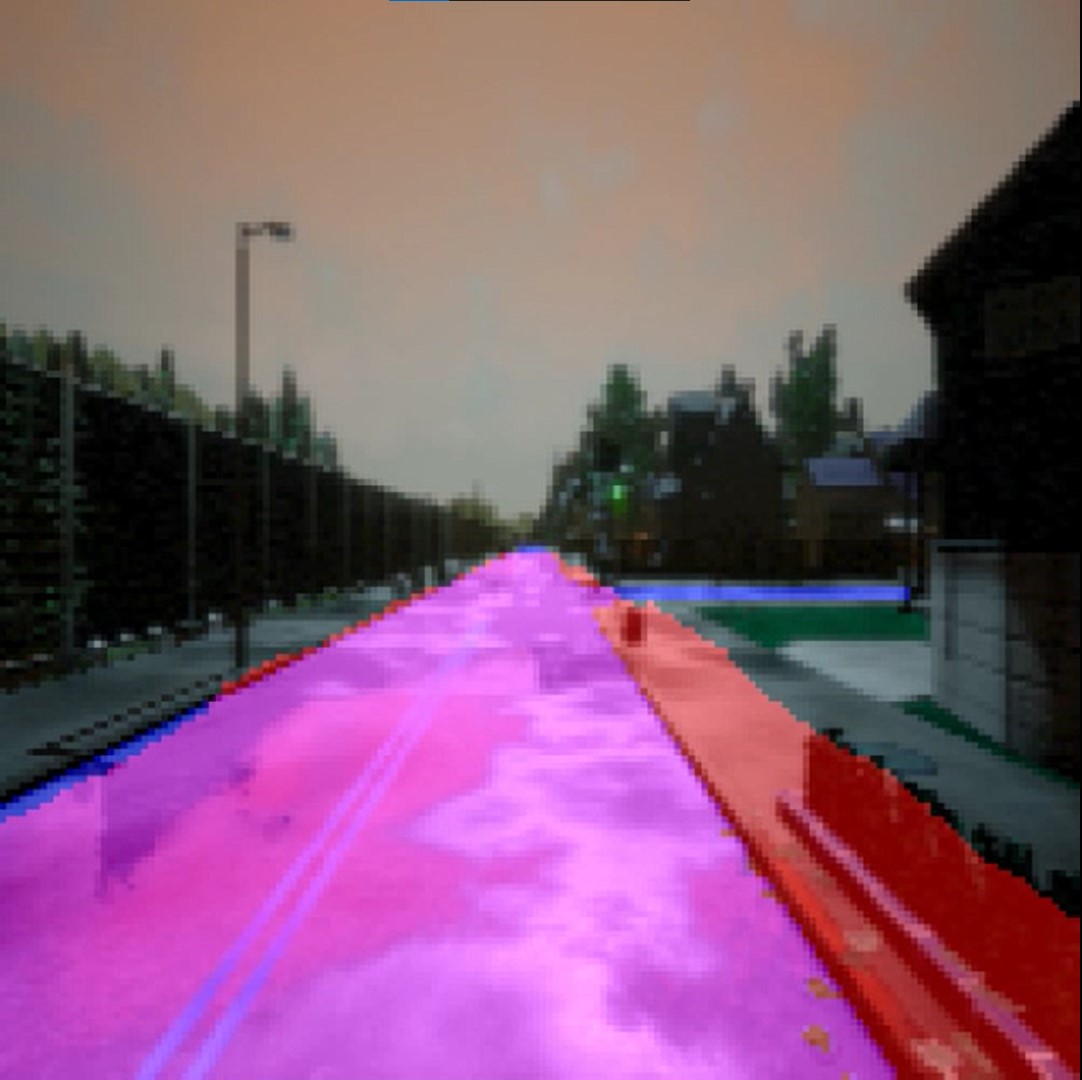}
        \caption{} \label{batch}
      \end{subfigure}
    \hspace*{0.1 cm}   % maximize separation between the subfigures
      \begin{subfigure}{0.32\linewidth}
        \includegraphics[width=\linewidth,height=3cm]{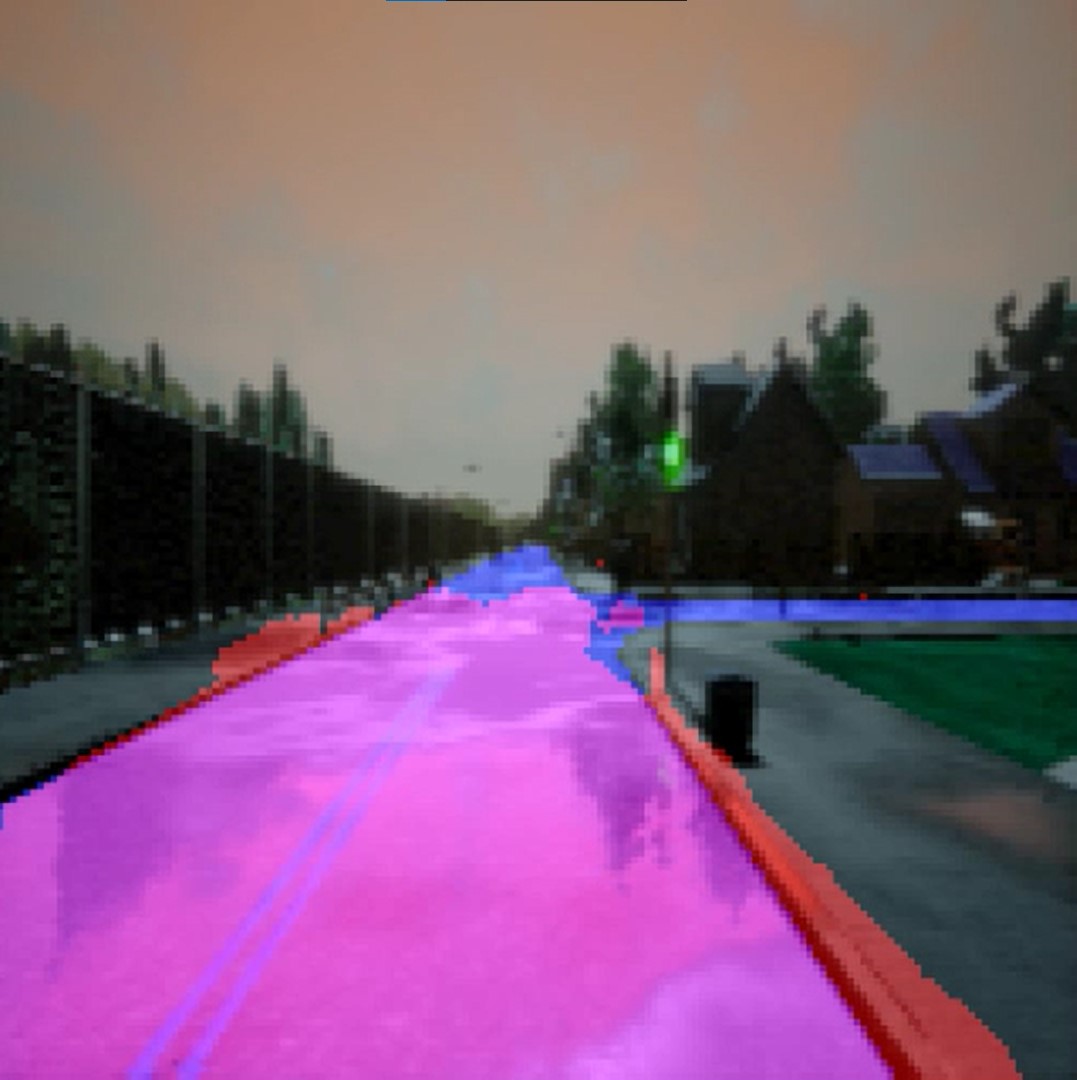}
        \caption{} \label{sequential}
      \end{subfigure}
    \caption{(a) Batched Interleaved (b) Sequential Interleaved} \label{fig:1}
\end{figure}

Comparing results between different models, we see that the Vanilla ResNet models and interleaved models perform similarly at each given frame. We notice that occasionally the prediction moves into the pavement as well. However, it is evident from the videos that the temporal consistency of the interleaved models is far superior than the vanilla models which simply treat each frame as an independent prediction.

In the videos, we notice that predictions immediately after clearing the weights of the LSTM are quite poor. This can be attributed to the loss being computed only for the last frame in a given sequence, which does not enforce the first few frames to have good predictions. We solve this problem by passing the features through the ConvLSTM before the features are propagated through the randomized feature extractor.

We also notice that the predictions are the worse when there are sharp/sudden changes in the video. This is to be expected since we are trying to use the temporal consistency of the video to make predictions.

Videos corresponding to each model can be found here: \href{https://drive.google.com/drive/u/0/folders/1D292uHyGZveTPtKJNERpBiiammGDYJb0}{Link}

\subsection{Quantitative Results}
\begin{table}[H]
\hspace{0.6cm}
\begin{tabular}{|l|c|c|c|}
\hline
\multicolumn{1}{|c|}{\textbf{Name}} & \textbf{Strategy}      & \textbf{Avg IoU} & \textbf{Avg FPS} \\ \hline
Resnet 18                           & Vanilla                & 0.852            & \textbf{155.4}   \\ \hline
Resnet 101                          & Vanilla                & \textbf{0.915}   & 46.18            \\ \hline
                                    &                        &                  &                  \\ \hline
Batched interleaved                 & Randn \textgreater 0.9 & 0.878            & 132.58           \\ \hline
Batched interleaved                 & 1 in 6                 & 0.877            & 135.73           \\ \hline
Batched interleaved                 & 1 in 10                & \textbf{0.88}    & \textbf{146.02}  \\ \hline
Batched interleaved                 & 1 in 12                & 0.879            & 147.92           \\ \hline
                                    &                        & \textbf{}        & \textbf{}        \\ \hline
Sequential Interleaved              & Randn \textgreater 0.7 & 0.872            & 117.91           \\ \hline
Sequential Interleaved              & Randn \textgreater 0.8 & \textbf{0.872}   & 132.12           \\ \hline
Sequential Interleaved              & Randn \textgreater 0.9 & 0.871            & \textbf{146.53}  \\ \hline
Sequential Interleaved              & 1 in 6                 & \textbf{0.871}   & 132.42           \\ \hline
Sequential Interleaved              & 1 in 10                & 0.872            & \textbf{141.31}  \\ \hline
Sequential Interleaved              & 1 in 12                & 0.868            & 139.64           \\ \hline
\end{tabular}
\caption{Avg IoU and Avg FPS of Various Trained Models}
\end{table}

As seen from the table, we notice that the Vanilla ResNet18 performs the fastest, whereas the vanilla Resnet 101 is the most accurate. However, the batched interleaved and the sequqential interleaved have IOUs almost similar to Resnet 101 with the FPS achieved is nearly equal to that of vanilla Resnet18. 

We notice that the batched interleaved model performs better than the sequential interleaved model. 

All of the experiments were performed on a laptop on an NVIDIA GTX 1650 GPU, and AMD Ryzen 4600H CPU with 8GB RAM.

\section{Novelty}

Previous approaches to road detection employ methods that treat frames without having any temporal context. To the best of our knowledge, this is the first work that utilizes an interleaved suite of feature extractors for the problem of road detection. Apart from the interleaved model, we have also implemented the following, but have not included them in the comparison due to their weak performance:

\begin{itemize}
    \item DeepLabv3-Mobilenet based road segmentation
    \item Direct Convolutional LSTM based road segmentation
    \item Upsampling Resnet18 features to Resnet101 features
        \begin{itemize}
            \item Here we try to replicate the features extracted by the Resnet101 by appending extra layers at the end of the ResNet18 model and upsampling the feature vector size.
            \item By essentially allowing the ResNet18 to behave similar to the ResNet101, we hypothesize that the predictions would improve. However, we were unable to train the model completely due to time and resource constraints.
        \end{itemize}
\end{itemize}

\section{Observations \& Conclusion}

We summarize some key observations as follows:
\begin{itemize}
    \item The interleaved model performs almost 2.5x as fast as the slowest extractor while having a performance that averages the slow and the fast extractors.
    \item There is a high temporal consistency between successive predictions due to the LSTM cell present in the architecture.
    \item The strategy employed to use a particular feature extractor affects the performance of the interleaved model. Although there is not a very wide variation in the performance, we can optimize the performance by trying to clear the weights and using the larger extractor depending on changes in the frames themselves.
    \item If the weights are not cleared at appropriate times, the prediction of the lanes get accumulated and lead to poor predictions.
\end{itemize}

We choose pretrained ResNet models as they dont have to be fully trained on a large database from scratch. Since the ResNets are large models, training them from scratch will take a lot of time and processing power. Hence, in the future, we can obtain better results by training the model specifically for the task of segmentation.

In the future, different strategies can be implemented for interleaving. Furthermore, we can explore the possibility of using reinforcement learning based policies to choose the extractors to alleviate the problem of slowly changing lanes / fast changing lanes.

The number of LSTM cells can also be increased to improve the temporal consistency.

\section{Links}

\begin{enumerate}
    \item Dataset : \href{https://drive.google.com/drive/u/0/folders/1lNtkBCYgI6VG5arNmLh1cVMz128XWzUT}{Link}
    \item Colab File : \href{https://colab.research.google.com/drive/1q8a1p3qmfJJrIqRpOlvwB1qCiVTxalXW?usp=sharing}{Link}
    \item Videos of Inference : \href{https://drive.google.com/drive/u/0/folders/1D292uHyGZveTPtKJNERpBiiammGDYJb0}{Link}
    \item Trained Models : \href{https://drive.google.com/drive/u/0/folders/18U3wXehqxttqc6Cl89lLfsLFkk-Xl3c9}{Link}
    \item Github Repo: 
    \href{https://github.com/praveenVnktsh/Fast-Road-Detection}{Link}
\end{enumerate}

\bibliographystyle{acm}
\bibliography{ref}

\begin{thebibliography}{10}

\bibitem{noauthor_microsoft_nodate}
Microsoft {COCO}: {Common} {Objects} in {Context} {\textbar} {SpringerLink}.

\bibitem{noauthor_pascal_nodate}
The {Pascal} {Visual} {Object} {Classes} ({VOC}) {Challenge} {\textbar}
  {International} {Journal} of {Computer} {Vision}.

\bibitem{deng2009imagenet}
{\sc Deng, J., Dong, W., Socher, R., Li, L.-J., Li, K., and Fei-Fei, L.}
\newblock Imagenet: A large-scale hierarchical image database.
\newblock In {\em 2009 IEEE conference on computer vision and pattern
  recognition\/} (2009), Ieee, pp.~248--255.

\bibitem{dosovitskiy_carla_2017}
{\sc Dosovitskiy, A., Ros, G., Codevilla, F., Lopez, A., and Koltun, V.}
\newblock {CARLA}: {An} {Open} {Urban} {Driving} {Simulator}.
\newblock {\em arXiv:1711.03938 [cs]\/} (Nov. 2017).
\newblock arXiv: 1711.03938.

\bibitem{kim}
{\sc Kim, Y.}
\newblock Two-step recurnet - younghyun.

\bibitem{kolesnikov_big_2020}
{\sc Kolesnikov, A., Beyer, L., Zhai, X., Puigcerver, J., Yung, J., Gelly, S.,
  and Houlsby, N.}
\newblock Big {Transfer} ({BiT}): {General} {Visual} {Representation}
  {Learning}.
\newblock {\em arXiv:1912.11370 [cs]\/} (May 2020).
\newblock arXiv: 1912.11370 version: 3.

\bibitem{liu_looking_2019}
{\sc Liu, M., Zhu, M., White, M., Li, Y., and Kalenichenko, D.}
\newblock Looking {Fast} and {Slow}: {Memory}-{Guided} {Mobile} {Video}
  {Object} {Detection}.
\newblock {\em arXiv:1903.10172 [cs]\/} (Mar. 2019).
\newblock arXiv: 1903.10172.

\bibitem{rong_lgsvl_2020}
{\sc Rong, G., Shin, B.~H., Tabatabaee, H., Lu, Q., Lemke, S., Možeiko, M.,
  Boise, E., Uhm, G., Gerow, M., Mehta, S., Agafonov, E., Kim, T.~H., Sterner,
  E., Ushiroda, K., Reyes, M., Zelenkovsky, D., and Kim, S.}
\newblock {LGSVL} {Simulator}: {A} {High} {Fidelity} {Simulator} for
  {Autonomous} {Driving}.
\newblock {\em arXiv:2005.03778 [cs, eess]\/} (June 2020).
\newblock arXiv: 2005.03778.

\bibitem{shah_airsim_2017}
{\sc Shah, S., Dey, D., Lovett, C., and Kapoor, A.}
\newblock {AirSim}: {High}-{Fidelity} {Visual} and {Physical} {Simulation} for
  {Autonomous} {Vehicles}.
\newblock {\em arXiv:1705.05065 [cs]\/} (July 2017).
\newblock arXiv: 1705.05065.

\bibitem{shelhamer_fully_2017}
{\sc Shelhamer, E., Long, J., and Darrell, T.}
\newblock Fully {Convolutional} {Networks} for {Semantic} {Segmentation}.
\newblock {\em IEEE transactions on pattern analysis and machine intelligence
  39}, 4 (Apr. 2017), 640--651.

\bibitem{xavier_2019}
{\sc Xavier, A.}
\newblock An introduction to convlstm, Apr 2019.

\bibitem{yu_multi-scale_2016}
{\sc Yu, F., and Koltun, V.}
\newblock Multi-{Scale} {Context} {Aggregation} by {Dilated} {Convolutions}.
\newblock {\em arXiv:1511.07122 [cs]\/} (Apr. 2016).
\newblock arXiv: 1511.07122.

\end{thebibliography}

\end{document}